\documentclass[11pt,a4paper]{article}
\usepackage[hyperref]{acl2017}
\usepackage{times}
\usepackage{latexsym}
\usepackage{multirow}
\usepackage{url}
\usepackage{subfig}
\usepackage{graphicx}
\usepackage{framed}
\aclfinalcopy 

\title{The Role of Conversation Context for Sarcasm Detection \\
in Online Interactions}

\author{
 Debanjan Ghosh$^{\mathsection}$ Alexander Richard Fabbri$^{\dagger}$ Smaranda Muresan$^{\ddagger}$\\
  $^{\mathsection}$School of Communication  Information, Rutgers University, NJ, USA \\
  $^{\dagger}$Department of Computer Science, Columbia University, NY, USA \\
  $^{\ddagger}$Data Science Institute, Columbia University, NY, USA  \\
  {\fontsize{10}{1em}{\tt debanjan.ghosh@rutgers.edu, \{arf2145,smara@columbia.edu\}}}}

\date{}

\begin{document}
\maketitle
\begin{abstract}
Computational models for sarcasm detection have often relied on the content of utterances in isolation. However, speaker's sarcastic intent is not always obvious without additional context. Focusing on social media discussions, we investigate two issues: (1) does modeling of conversation context help in sarcasm detection and (2) can we understand what part of conversation context triggered the sarcastic reply. To address the first issue, we investigate several types of Long Short-Term Memory (LSTM) networks that can model both the conversation context and the sarcastic response.\footnote{We use response and reply interchangeably.} We show that the conditional LSTM network \cite{rocktaschel2015reasoning} and LSTM networks with sentence level attention on context and response outperform the LSTM model that reads only the response. To address the second issue, we present a qualitative analysis of attention weights produced by the LSTM models with attention and discuss the results compared with human performance on the task.  
\end{abstract}
\section{Introduction} \label{intro}
It has been argued that sarcasm, or verbal irony, is a type of interactional phenomenon with specific perlocutionary effects on the hearer \cite{haverkate1990speech}, such as to break their pattern of expectation. Thus, to be able to detect speakers' sarcastic intent it is necessary (even if maybe not sufficient) to consider their utterances in the larger conversation context. Consider the Twitter conversation example
in Table \ref{table:ex}. Without the context of UserA's
statement, the sarcastic intent of UserB's response might not be detected. 

Most computational models for sarcasm detection have considered utterances in isolation  \cite{davidov2010,gonzalez,liebrecht2013perfect,riloff,maynard2014cares,joshi2015,
ghoshguomuresan2015EMNLP,joshi2016word,ghosh2016fracking}. In many instances, even humans have difficulty in recognizing sarcastic intent when considering an utterance in isolation \cite{wallace2014humans}. 

\begin{table}
\begin{tabular}{ |l|p{5cm}| } 
\hline
\multicolumn{1}{|c|}{Platform} & 
\multicolumn{1}{c|}{Context-Reply pair} \\
\hline
\multirow{2}{*}{Twiter} & {\bf userA:} plane window shades are open \dots so that people can see if there is fire. \\ & {\bf userB}: @UserA one more reason to feel really great. \\ \hline
\multirow{2}{4em}{Discussion Forum} & {\bf userC:}  see for yourselves. The fact remains that in the caribbean, poverty and crime was near nil. Everyone was self-sufficient and contented with the standard of life.  there were no huge social gaps.   \\
& {\bf userD:} Are you kidding me?! You think that Caribbean countries are ``content?!'' Maybe you should wander off the beach sometime and see for yourself.\\
\hline
\end{tabular}
\caption{Sample Context/Reply pairs from two social media platforms}
\label{table:ex}
\end{table}

In this paper, we investigate the role of  \textit{conversation context} in detecting sarcasm in social media discussions (Twitter conversations and discussion forums). Table \ref{table:ex} shows some examples of sarcastic replies taken from two media platforms (userB and userD's posts, respectively) and a minimum unit of conversation context given by the prior turn (userA and userC's posts, respectively).

We address two specific issues: (1) does modeling of conversation context help in sarcasm detection and (2) can we understand what part of conversation context triggered the sarcastic reply (e.g., which sentence(s) from userC's comment triggered userD's sarcastic reply). To address the first issue, we investigate both SVM models with linguistically-motivated discrete features and several types of Long Short-Term Memory (LSTM) networks 
\cite{hochreiter1997long} that can model both the  context and the sarcastic reply (Section \ref{section:experiment}). We show that the conditional LSTM network \cite{rocktaschel2015reasoning} and LSTM networks with sentence level attention on context and reply outperform the LSTM model that reads only the reply (Section \ref{section:results}). 
To address the second issue, we present a qualitative analysis of attention weights produced by the LSTM models with attention, and discuss the results compared with human performance on the task (Section \ref{section:qualitative}). We make all datasets and code available.\footnote{https://github.com/debanjanghosh/sarcasm\_context}

\section{Data} \label{section:corpus}

One goal of our investigation is to comparatively study two types of social media platforms that have been considered individually for sarcasm detection:  discussion forums and Twitter. We first discuss the two  datasets and then point out some differences between them that could impact results and modeling choices. 

\paragraph{Discussion Forums.}  
\newcite{orabycreating}
have introduced the Sarcasm Corpus V2, a subset of the Internet Argument Corpus that consists of discussion forum data. This corpus consists of sarcastic responses and their context (quotes to which the posts are replies to). The annotation of sarcastic vs. non-sarcastic replies was done using crowdsourcing, where annotators were asked to label a reply as sarcastic if any part of the reply contained sarcasm (thus annotation is done at the reply/comment level and not sentence level). The final gold sarcastic label was assigned only if a majority of the annotators labeled the reply as sarcastic. Although the dataset described by \newcite{orabycreating} consists of 9,400 post, only 50\% (4,692 altogether; balanced between sarcastic and non-sarcastic categories)  of that corpus is currently available for research.\footnote{This reduction in the training size will have obvious effects in the classification performance.}


An example from this dataset is given in Table \ref{table:ex}, where userD's reply has been labeled as sarcastic by annotators, in the context of userC's post/comment.

\paragraph{Twitter:}
To collect sarcastic and non-sarcastic tweets, we adopt the methodology proposed in related work  \cite{gonzalez,riloff,bamman2015contextualized,muresanjasist2016}. The sarcastic tweets were collected using hashtags such as, \emph{\#sarcasm}, \emph{\#sarcastic}, \emph{\#irony}, while the non-sarcastic tweets were the ones that do not contain these hashtags, but they might contain sentiment hashtags such as  \emph{\#happy}, \emph{\#love}, \emph{\#sad}, \emph{\#hate}. We exclude the retweets, duplicates, quotes, tweets that contain only hashtags and URLs or are shorter than three words. Also, we eliminate all tweets where the hashtags of interest were not positioned at the very end of the message. Thus, we removed utterances such as ``\#sarcasm is something that I love''. To built the conversation context, for each sarcastic and non-sarcastic utterance we used the ``reply to status'' parameter in the tweet to determine whether it was in reply to a previous tweet: if so, we downloaded the last tweet (i.e., ``local conversation context'') to which the original tweet was replying to \cite{bamman2015contextualized}. In addition, we also collected the entire threaded conversation when available \cite{wang2015twitter}. Although we have collected over 200K tweets in the first step, around 13\% of them were a reply to another tweet and thus our final Twitter conversations set contains 25,991 instances (12,215 instances for sarcastic class and 13,776 instances for the non-sarcastic class). We observe that 30\% of the tweets have more than one tweet in the conversation context. 

There are two main differences between these two datasets that need to be acknowledged. First, discussion forum posts are much longer than Twitter messages. Second, the way the gold labels for the sarcastic class are obtained is different. In the discussion forum dataset the gold label is obtained via crowdsourcing, thus the gold label emphasizes whether the sarcastic intent is \emph{perceived} by hearers (we do not know if the speaker intended to be sarcastic or not).  In Twitter dataset the gold label is given directly by the \#hashtag the speaker used, signaling clearly the speaker's sarcastic intent. A third difference should be made: the size of the forum dataset is much smaller than the size of the Twitter dataset.

\section{Computational Models and Experimental Setup} \label{section:experiment}

To assess the effect of conversation context ($c$) on labeling a reply ($r$) as sarcastic or not sarcastic, we consider two binary classification tasks. We refer to sarcastic instances as $S$ and non-sarcastic instances as $NS$. 
In the first task, classification is performed using the reply in isolation ($S^{r}$ vs. $NS^{r}$ task). In the second, the classification considers both the reply and its context ($S^{c+r}$ vs. $NS^{c+r}$ task). 
We experiment with two types of computational models: Support Vector Machines (SVM) with linguistically-motivated discrete features (used as baseline; SVM$_{bl}$), and approaches using distributed representations. For the latter we use the Long short-term Memory (LSTM) Networks \cite{hochreiter1997long} that have been shown to be successful in various NLP tasks, such as constituency parsing \cite{vinyals2015grammar}, language modeling \cite{zaremba2014recurrent}, machine translation \cite{sutskever2014sequence} and textual entailment \cite{bowman2015large,rocktaschel2015reasoning,parikh2016decomposable}. We present these models in the next subsections.

\subsection{SVM with discrete features (SVM$_{bl}$)}  

For features, we used n-grams, lexicon-based features, and sarcasm indicators that are commonly used in the existing sarcasm detection approaches \cite{tchokni2014,gonzalez,riloff,joshi2015,ghoshguomuresan2015EMNLP,muresanjasist2016}. Below is a short description of the features.
\begin{itemize}
\item \textbf{BoW:} Features are derived from unigram, bigram, and trigram representation of words.
\item \textbf{Sentiment and Pragmatic features:} We use the Linguistic Inquiry and Word Count (LIWC) lexicon \cite{pennebaker2001} to identify the pragmatic features. Each category in this dictionary is treated as a separate feature 
and we define a Boolean feature that indicates if a context or a reply contains a LIWC category. Two sentiment lexicons are also used to model the utterance sentiment: ``MPQA'' \cite{wilson2005recognizing} and ``Opinion Lexicon'' \cite{hu2004mining}. To capture sentiment, we count the number of positive and negative sentiment tokens, negations, and use a boolean feature that represents whether a reply contains both positive and negative sentiment tokens. For the $S^{c+r}$ vs. $NS^{c+r}$ classification task, we check whether the reply $r$ has a different sentiment than the context $c$ (similar to \newcite{joshi2015}). Given that sarcastic utterances often contain a positive sentiment towards a negative situation, we hypothesize that this feature will capture this type of sentiment incongruity.
\item \textbf{Sarcasm Indicators:} \newcite{burgers2012verbal} introduce a set of sarcasm indicators that explicitly signal if an utterance is sarcastic. We use \textit{morpho-syntactic} features such as interjections (e.g., ``uh'', ``oh'', ``yeah''), tag questions (e.g., ``is not it?'', ``don't they''), exclamation marks (e.g., ``!'', ``?''); \textit{typographic} features such as capitalization of words, quotation marks, emoticons; \textit{tropes} such as superlative and intensifiers words (e.g., ``greatest'', ``best'', ``really") that often occur in sarcastic utterances \cite{camp2011}.
\end{itemize}                                                    

When building the features, we lowercased the utterances, except the words where all the characters are uppercased (i.e., we did not lowercased ``GREAT'', ``SO'', and ``WONDERFUL'' in ``GREAT i'm SO happy;  shattered phone on this WONDERFUL day!!!''). Tokenization is conducted via CMU's Tweeboparser \cite{gimpel2011part}. For the discussion forum dataset we use the NLTK tool \cite{bird2009natural} for sentence boundary detection and tokenization. We used libSVM toolkit with Linear Kernel \cite{svm} with weights inversely proportional to the number of instances in each class.

\subsection{Long Short-Term Memory Networks}

LSTMs are a type of recurrent neural networks (RNNs) able to learn long-term dependencies \cite{hochreiter1997long}. Recently, LSTMs have been shown to be effective in Natural Language Inference (NLI) research, where the task is to establish the \emph{relationship} between multiple inputs (i.e., a pair of premise and hypothesis as in the case of Recognizing Textual Entailment task \cite{bowman2015large,rocktaschel2015reasoning,parikh2016decomposable}). Since our goal is to explore the role of contextual information (our \textit{first input}) for recognizing whether the reply (our \textit{second input}) is sarcastic or not, we argue that using LSTM networks that read the context and reply are a natural modeling choice. 

\paragraph{Attention-based LSTM Networks:}
Attentive neural networks have been shown to perform well on a variety of NLP tasks \cite{yang2016hierarchical,yin2015abcnn,xu2015show}. Using attention-based LSTM will accomplish two goals: (1) test whether they achieve higher performance than simple LSTM models and (2) use the attention weights produced by the LSTM models to perform a qualitative analysis to determine which portions of context triggers the sarcastic reply. 

Although \newcite{yang2016hierarchical} have included two levels of attention mechanisms -- one at the word level and another at the sentence level -- we primarily focus on sentence level attention for two specific reasons. First, sentence level attentions can show the exact sentence in the context that is most informative to trigger sarcasm. In the discussion forum dataset, context posts are usually three or four sentences long and it could be helpful to identify the exact text that triggers the sarcastic reply. Second, attention over both the words and sentences seek to learn a large number of model parameters and given the moderate size of the discussion forum corpus they might overfit.   
For tweets, we treat each individual tweet as a sentence. The majority of tweets consist of a single sentence and even if there are multiple sentences in a tweet, often one sentence contains only hashtags, URLs, and emoticons making them uninformative if treated in isolation.

\begin{figure}[t]
\begin{framed}
\includegraphics[width=7cm]{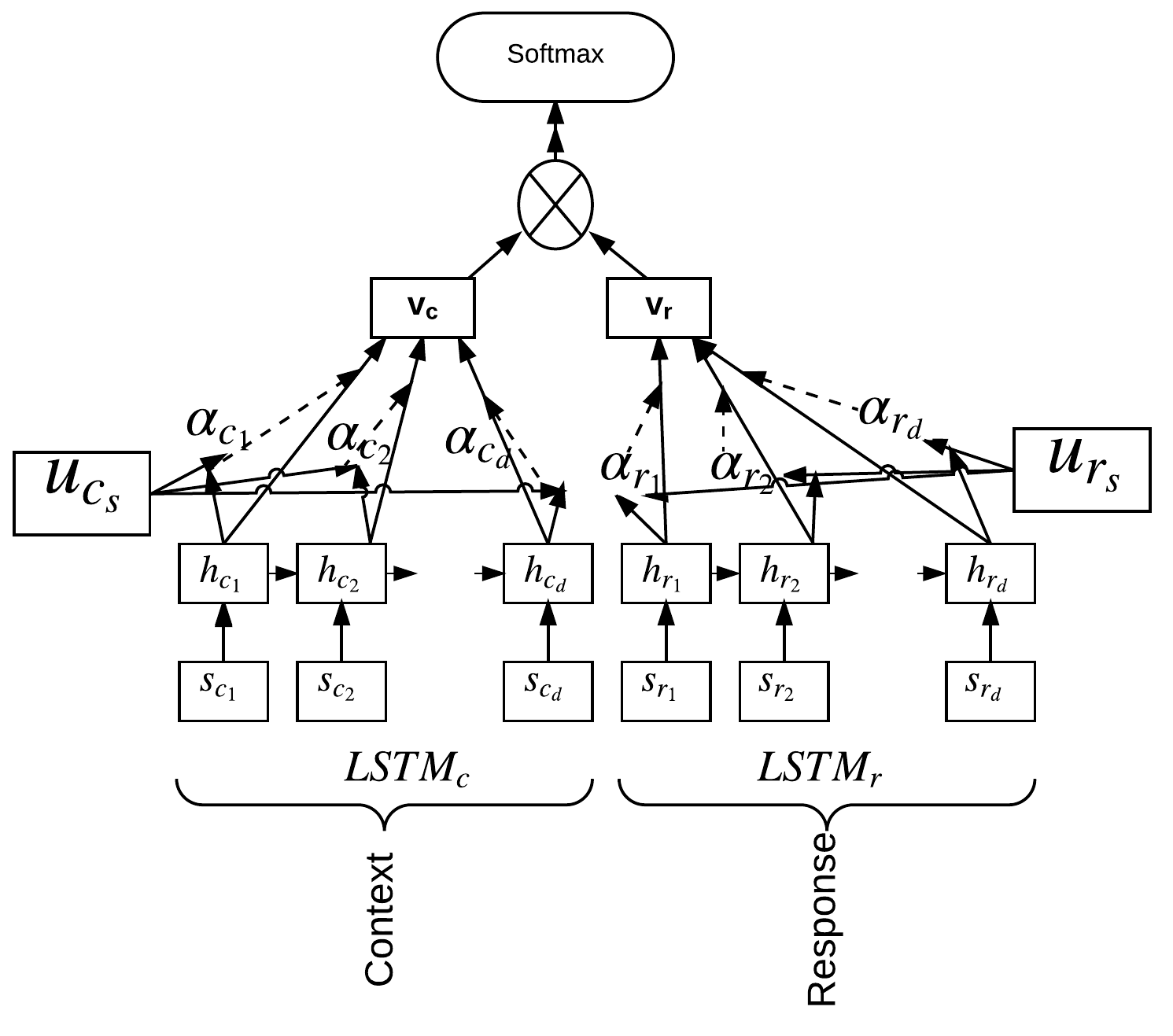}
\end{framed}
\caption{Sentence-level Attention Network for Context and Reply. Figure is inspired by \newcite{yang2016hierarchical}}
\label{figure:model}
\end{figure}

Figure \ref{figure:model} shows the high-level structure of the model. The context (left) is read by an LSTM ($LSTM_{c}$) whereas the response (right) is read by another LSTM ($LSTM_{r}$). We represent each sentence by the average of its word embeddings. 

Let the context $c$ contain $d$ sentences and each sentence $s_{c_{i}}$ contain $T_{c_{i}}$ words. Similar to the notation of \newcite{yang2016hierarchical}, we first feed the sentence annotation $h_{c_{i}}$ through a one layer MLP to get $u_{c_{i}}$ as a hidden representation of $h_{c_{i}}$, then we weight the sentence $u_{c_{i}}$ by measuring similarity with a sentence level context vector $u_{c_{s}}$. This gives a normalized importance weight $\alpha_{c_{i}}$ through a softmax function. $v_c$ is the vector that summarize all the information of sentences in the context ($LSTM_{c}$).
\begin{equation}
v_c = \sum_{i\in[1,d]} \alpha_{i_{c}}h_{i_{c}}
\end{equation}
where attention is calculated as:
\begin{equation}
\alpha_{i_{c}} = \frac{\exp(u_{c_{i}}^Tu_{c_{s}} )}{\sum_{i\in[1,d]}\exp(u_{c_{i}}^Tu_{c_{s}} )}
\end{equation}

Likewise we compute $v_r$ for the response $r$ via $LSTM_{r}$ (similar to eq. 1 and 2; also shown in Figure \ref{figure:model}). Finally, we concatenate the vector $v_c$ and $v_r$ from the two LSTMs for the final softmax decision (i.e., predicting the $S$ or $NS$ class).

We also experiment with both word and sentence level attentions in a hierarchical fashion similarly to the approach proposed by \newcite{yang2016hierarchical}. As we show in Section \ref{section:results} however, we achieve best performance for both datasets using just the sentence-level attention. 

\paragraph{Conditional LSTM Networks:}
We also experiment with the \emph{conditional encoding} model as introduced by  \newcite{rocktaschel2015reasoning} for the task of recognizing textual entailment. In this architecture,  two separate LSTMs are used -- $LSTM_{c}$ and $LSTM_{r}$ -- similar to the previous architecture without any attention, but for $LSTM_{r}$, its memory state is initialized with the last cell state of $LSTM_{c}$. In other words, $LSTM_{r}$ is conditioned on the representation of $LSTM_{c}$ that is built on the context. 

\paragraph{\textbf{Parameters and pre-trained word vectors.}}
For both discussion forum and Twitter, we split randomly the corpus into training (80\%), development (10\%), and test (10\%), maintaining the same distribution of sarcastic vs. non-sarcastic data in training, development and test. For Twitter we used the skip-gram word-embeddings (100-dimension) used in \cite{ghoshguomuresan2015EMNLP} that was built using over 2.5 million tweets.\footnote{https://github.com/debanjanghosh/sarcasm\_wsd} For discussion forums, we use the standard Google n-gram $word2vec$ pre-trained model (300-dimension) \cite{mikolov2013efficient}. We do not optimize the word embedding during training. Out-of-vocabulary words in the training set are randomly initialized via sampling values uniformly from (-0.05,0.05). We use the development data to tune the parameters and selected dropout rate of 0.5 (from [.25,0.5, 0.75]), $L_{2}$ regularization strength and evaluate only that configuration on the test set. For both datasets mini-batch size of 16 is employed. 
\section{Results and Discussion} \label{section:results} 

\begin{table*} [t]
\centering
\begin{tabular}{||c|c|c|c|c|c|c||}
\hline
\multirow {2}{*}{Experiment} & \multicolumn{3}{c}{$S$} &  \multicolumn{3}{c||}{$NS$} \\
& P & R & F1 & P & R & F1  \\
\hline
{SVM$_{bl}^{r}$}  & 65.55  & 66.67  &  66.10   &  66.10  &  64.96 &  65.52   \\
\hline
{SVM$_{bl}^{c+r}$}  & 63.32  & 61.97  &  62.63   &  62.77  &  64.10 &  63.5   \\
\hline
{LSTM$^{r}$}  & 67.90  & 66.23   & 67.1  & 67.08   & \textbf{68.80}  & 67.93    \\
\hline
{LSTM$^{c}$+LSTM$^{r}$}  & 66.19 & 79.49 & 72.23 & 74.33 & 59.40 & 66.03  \\
\hline
{LSTM$^{conditional}$} & \textbf{70.03} & 76.92 & \textbf{73.32} &  74.41 & 67.10 & \textbf{70.56} \\ \hline
{LSTM$^{r_{a_{s}}}$}  & 69.45 & {70.94} &  {70.19} & {70.30} & 68.80 & 69.45 \\
{LSTM$^{c_{a_{s}}}$+LSTM$^{r_{a_{s}}}$}  & 66.90 & \textbf{82.05} & \textbf{73.70} & \textbf{76.80} & 59.40 & 66.99 \\
{LSTM$^{c_{a_{w+s}}}$+LSTM$^{r_{a_{w+s}}}$}  & 65.90 & 74.35 & 69.88 & 70.59 & 61.53 & 65.75 \\
\hline
\end{tabular}
\caption{Experimental results for the discussion forum dataset ({\bf bold} are best scores)}
\centering
\label{table:results2}
\end{table*}
We report Precision (P), Recall (R), and F1 scores on $S$ and $NS$ classes. SVM$_{bl}^{r}$ and SVM$_{bl}^{c+r}$ respectively represent the performance of the SVM model using discrete features when using only the reply and the reply together with context. LSTM$^{c_{a}}$ and LSTM$^{r_{a}}$ are the attention-based LSTM models of context and reply, where the $w$, $s$ and $w+s$ subscripts denote the word-level, sentence-level or word and sentence level attentions. LSTM$^{conditional}$ is the \emph{conditional encoding} model (no attention). 

\paragraph{Discussion Forums:} Table \ref{table:results2} shows the classification results on the discussion forum dataset. Although a vast majority of the context posts contain 3-4 sentences, around 100 context posts have more than ten sentences and thus we set a cutoff to a maximum of ten sentences for context modeling. For the reply $r$ we considered the entire reply. 

\begin{table*} [t] 
\centering
\begin{tabular}{||c|c|c|c|c|c|c||}
\hline
\multirow {2}{*}{Experiment} & \multicolumn{3}{c}{$S$} &  \multicolumn{3}{c||}{$NS$} \\
& P & R & F1 & P & R & F1  \\
\hline
{SVM$_{bl}^{r}$} &	64.20	&	64.95	&	64.57	&	69.0	&	68.30	&	68.7 \\
\hline
{SVM$_{bl}^{c+r}$} & 	65.64&		65.86 &		65.75	&	70.11 & 69.91	&	70.0 \\
\hline
{LSTM$^{r}$}	& 73.25	&	58.72	&	65.19	&	61.47	&	75.44	&	67.74 \\
\hline
{LSTM$^{c}$+LSTM$^{r}$}	 & 70.89	&	67.95	&	69.39 &		64.94	 &	68.03	&	66.45 \\ 
\hline
{LSTM$^{conditional}$} &	76.08	 &	\textbf{76.53}	&	\textbf{76.30}	&	\textbf{72.93}	&	72.44		& \textbf{72.68}  \\
\hline	
{LSTM$^{r_{a_{s}}}$} &	{76.00}	&	73.18	&	{74.56} 	&	70.52	&	{73.52}	&	{71.9} \\
{LSTM$^{c_{a_{s}}}$+LSTM$^{r_{a_{s}}}$} &	\textbf{77.25}	&	75.51	&	\textbf{76.36} 	&	72.65	&	\textbf{74.52}	&	\textbf{73.57} \\
{LSTM$^{c_{a_{w}}}$+LSTM$^{r_{a_{w}}}$} &	76.74	&	69.77	&	73.09	&	68.63	&	75.77	&	72.02 \\
{LSTM$^{c_{a_{w+s}}}$+LSTM$^{r_{a_{w+s}}}$} 	& 76.42	&	71.37	&	73.81	&	69.50	&	74.77	&	72.04 \\
\hline

\end{tabular}
\caption{Experimental results for Twitter dataset ({\bf bold} are best scores)}
\label{table:results1}
\end{table*}

The $SVM_{bl}$ models that are based on discrete features did not perform very well, and adding context actually hurt the performance. 
Regarding the performance of the neural network models, we observe that modeling context improves the performance using all types of LSTM architectures that read both context ($c$) and reply ($r$) (results are statistically significant when compared to LSTM$^r$). The highest performance when considering both the $S$ and $NS$ classes is achieved by the LSTM$^{conditional}$ model (73.32\% F1 for $S$ class and 70.56\% F1 for $NS$, showing a 6\% and 3\% improvement over LSTM$^r$ for $S$ and $NS$ classes, respectively). The LSTM model with sentence-level attentions on both context and reply ({LSTM$^{c_{a_{s}}}$+LSTM$^{r_{a_{s}}}$}) gives the best F1 score of 73.7\% for the $S$ class. For the $NS$ class, while we notice an improvement in precision we notice a drop in recall when compared to the LSTM model with sentence level attention only on reply (LSTM$^{r_{a_{s}}}$).  Remember that sentence-level attentions are based on average word embeddings. We also experimented with the hierarchical attention model where each sentence is represented by a \textit{weighted average} of its word embeddings. In this case, attentions are based on words and sentences and we follow the architecture of hierarchical attention network \cite{yang2016hierarchical}. We observe the performance (69.88\% F1 for $S$ category) deteriorates, probably due to the lack of enough training data. Since attention over both the words and sentences seek to learn a lot more model parameters, adding more training data will be helpful. With the full release of the Sarcasm Corpus used by \newcite{orabycreating}, we expect to achieve better accuracy for these models. 


\paragraph{Twitter:} Table \ref{table:results1} shows the results on the Twitter dataset. As for discussion forums, adding context using the SVM models does not show a statistically significant improvement.  
For the neural networks model, similar to the results on discussion forums, the LSTM models that read both context and reply outperform  the LSTM model that reads only the reply (LSTM$^{r}$). The best performing architectures are again the LSTM$^{conditional}$ and LSTM with sentence-level attentions ({LSTM$^{c_{a_{s}}}$+LSTM$^{r_{a_{s}}}$}). LSTM$^{conditional}$ model shows an improvement of 11\% F1 on the $S$ class and 4-5\%F1 on the $NS$ class, compared to LSTM$^{r}$. For the attention-based models, the improvement using context is smaller ($\sim$2\% F1). We kept the maximum length of context to the last five tweets in the conversation context, when available.
  We also conducted experiments with only word-level attentions, however, we obtain lower accuracy in comparison to sentence level attention models.


\subsection{Qualitative Analysis} \label{section:qualitative}


\newcite{wallace2014humans} showed that by providing  contextual information humans are able to identify sarcastic utterances which they were unable without the context. However, it will be useful to understand whether a specific \textit{part of the context} triggers the sarcastic reply.    

To begin to address this issue, we conducted a qualitative study to understand whether (a) human annotators are able to identify parts of context  that trigger the sarcastic reply and (b) attention weights are able to signal similar information. For (a) we designed a crowdsourcing experiment and for (b) we looked at the attention weights of the LSTM networks. 
Below is a short description of the crowdsourcing task.

\subsubsection{Crowdsourcing Experiment.} We designed an Amazon Mechanical Turk task (for brevity, MTurk) framed as follow: Given a pair of context $c$ and a sarcastic reply $r$ from the discussion forum dataset, identify one or more sentences in $c$ that may trigger the sarcastic reply $r$. Turkers could select one or more sentences from the context $c$, including the entire context. From the test data, we select examples with context length between three to seven sentences since for longer posts the task will be too complicated for the Turkers. 

We provided a definition of sarcasm and a few examples to the Turkers. We also explained how to carry out the task with the help of a few context/reply pairs. Each HIT contains only one task and five Turkers were allowed to attempt each HIT (a total of 85 HITS).  Turkers with reasonable quality (i.e., more than 95\% of acceptance rate with experience of over 8,000 HITs) were selected and paid seven cents per task. 

\subsubsection{Comparing Turkers' answers with attention models.}
\begin{figure*}[t]
\centering
\begin{framed}
 \subfloat[\label{fig:hmap1}]
 {\includegraphics[width=1.7in]{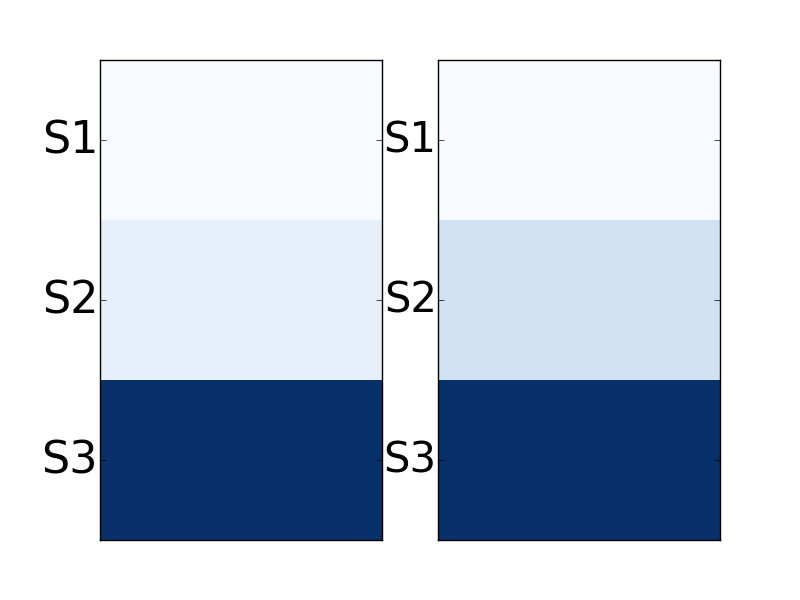}} 
 \subfloat[ \label{fig:hmap2}]
 {\includegraphics[width=1.7in]{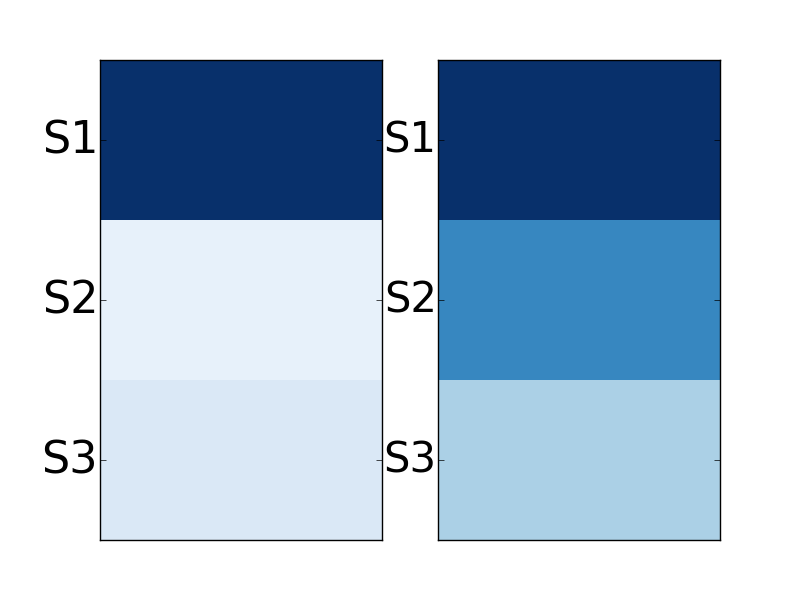}}
 \subfloat[\label{fig:hmap3}]
 {\includegraphics[width=1.7in]{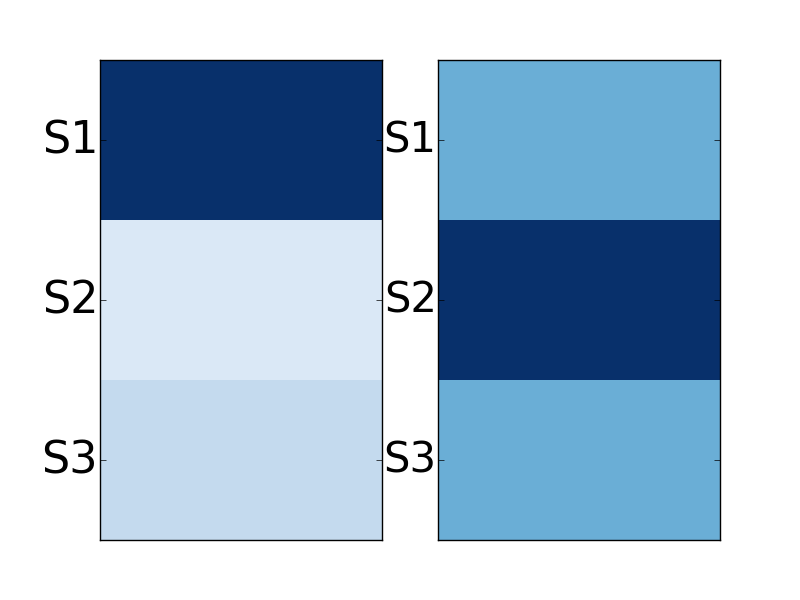}}
 \end{framed}
 \centering
 \caption{Context sentences that trigger sarcasm: LHS: \textit{attention weights}; RHS: \textit{Turkers' selections}}
 \label{figure:iachits}
\end{figure*}

We visualize and compare the sentence-level attention weights of the LSTM models on context with Turkers' annotations (Figure \ref{figure:iachits}). We first measure the overlap of Turkers choice with the attention weights. For the sentence-based attention model (i.e., LSTM$^{c_{a_{s}}}$+LSTM$^{r_{a_{s}}}$ model for the discussion forum), we selected the sentence with highest attention weight and matched it to the sentence selected by Turkers using majority voting. We found that 41\% of times the sentence with the highest attention weight is also the one picked by Turkers. 
Figure \ref{figure:iachits} shows side by side the heat maps of the attention weights of LSTM models (LHS) and Turkers' choices when picking up sentences from context that they thought triggered the sarcastic reply (RHS).


Here the obvious question that we need to answer is why these sentences are selected by the models (and humans). In the next section we conduct a qualitative analysis to try answering this question. 

\subsubsection{Interpretation of selected context via attention weights}

\begin{figure}[t]
\centering
\begin{framed}
 {\includegraphics[width=1.7in]{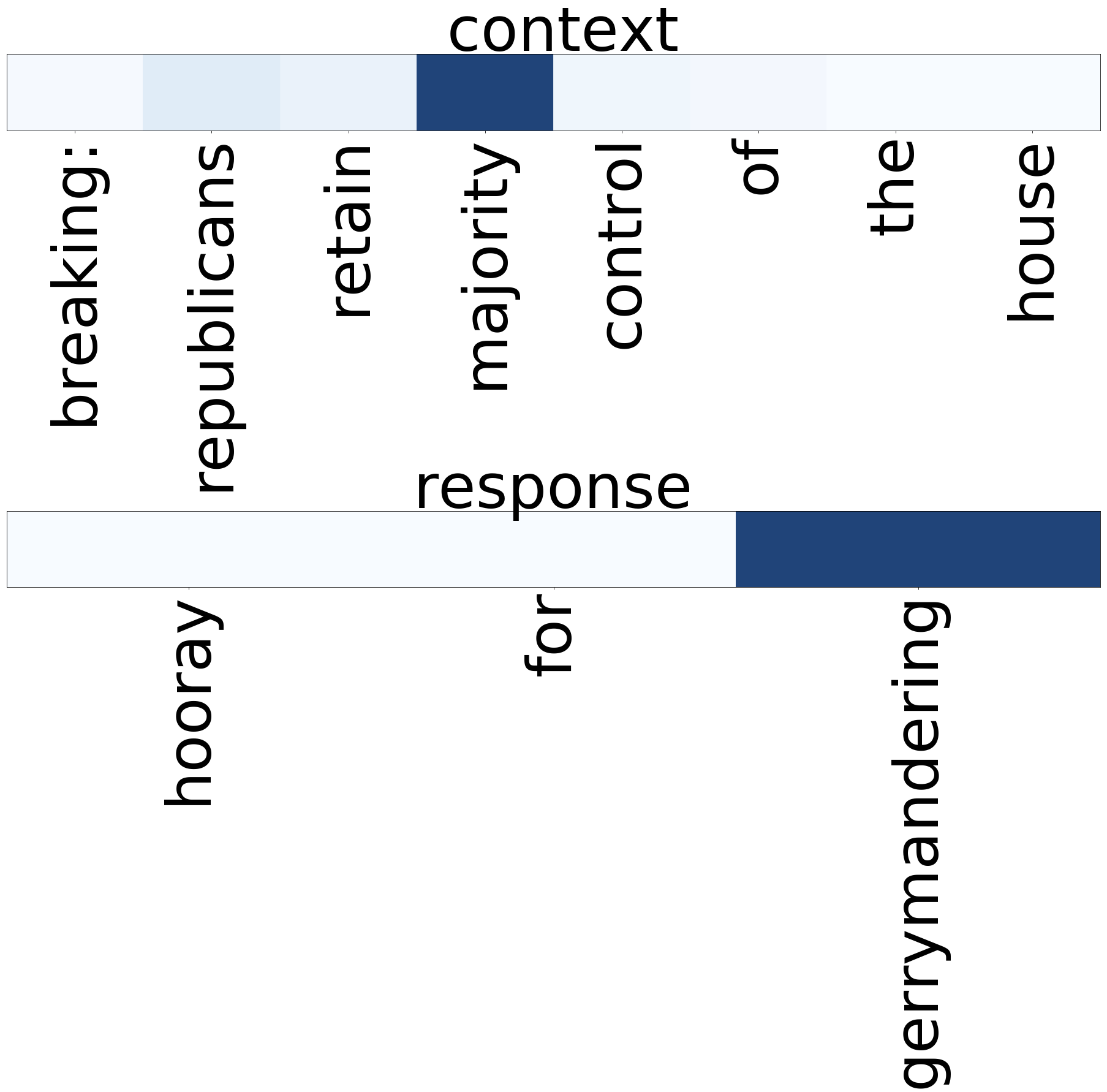}}
 \end{framed}
   \caption{Attention visualization of semantic coherence between $c$ and $r$}
  \label{figure:wordmap1}
\end{figure}

\paragraph{Semantic coherence between context and reply.}
Figure \ref{figure:iachits}(a) depicts a case where the context contains three sentences and the attention weights given to the sentences are similar to the Turkers' choice. Looking at this example it seems the model pays attention to output vectors that are semantically coherent between $c$ and $r$. The sarcastic response of this example contains a single sentence --
``\dots hold your tongue \dots in support of an anti-gay argument''. The context contains  the sentence S3 ``\dots I've held my tongue on this as long as I can''.  The attention-based LSTM architecture is learning the attention weights simultaneously for the context $c$ and the response $r$. Thus the model is showing contextual understanding by setting high weights to semantically coherent parts of the $c$ and $r$. 
In Figure \ref{figure:iachits}(b), attention weights is given to the most informative sentence --``rationally explain these creatures existence so recently in our human history if they were extinct for millions of years?''. Here, the sarcastic reply mocks  by claiming the author of the context is reading a lot more religious script (`` you're reading waaaaay too much into your precious bible''). We also observe similar behavior in Tweets (highest attention to words --\textit{retain} and \textit{gerrymadering} in context: ``breaking: \textit{republicans retain majority} control of house'' and reply: ``hooray for \textit{gerrymandering}'' (Figure \ref{figure:wordmap1}). 


 
\paragraph{Incongruity between context and reply}
The meaning incongruity is an inherent characteristic of irony and sarcasm and have been extensively studied in linguistics, philosophy, communication science \cite{grice1975syntax,attardo2000irony,burgers2012verbal} as well as recently in NLP \cite{riloff,joshi2015}. For instance, \newcite{riloff} pointed out that identifying the incongruity between \textit{positive} sentiment towards a \textit{negative} situation is a key characteristic of sarcasm detection in social media. We observe in discussion forums and in Tweets that the attention-based models have frequently identified sentences and words from $c$ and $r$ that are semantically incongruous (i.e., opposite sentiment words). For instance, in Figure \ref{figure:iachits}(c), the attention model has chosen sentence S1, which contains strong negative sentiment word (``disgusting sickening \dots''). Interestingly, in contrast, the attention model on the reply, has given the highest weight to sentence that contain opposite sentiment  (``I love you''). Thus, the model seems to learn the context incongruity of opposite sentiment for detecting sarcasm. However, it seems the Turkers prefer the second sentence S2 (``how can you tell a man that about his mum?'') as the most instructive sentence instead of the first sentence. Looking at the sarcastic reply we observe that the reply contains remarks about ``mothers'' and apparently that commonality assisted the Turkers to chose the second sentence.  

\begin{figure}[t]
\centering
\begin{framed}
 {\includegraphics[width=1.7in]{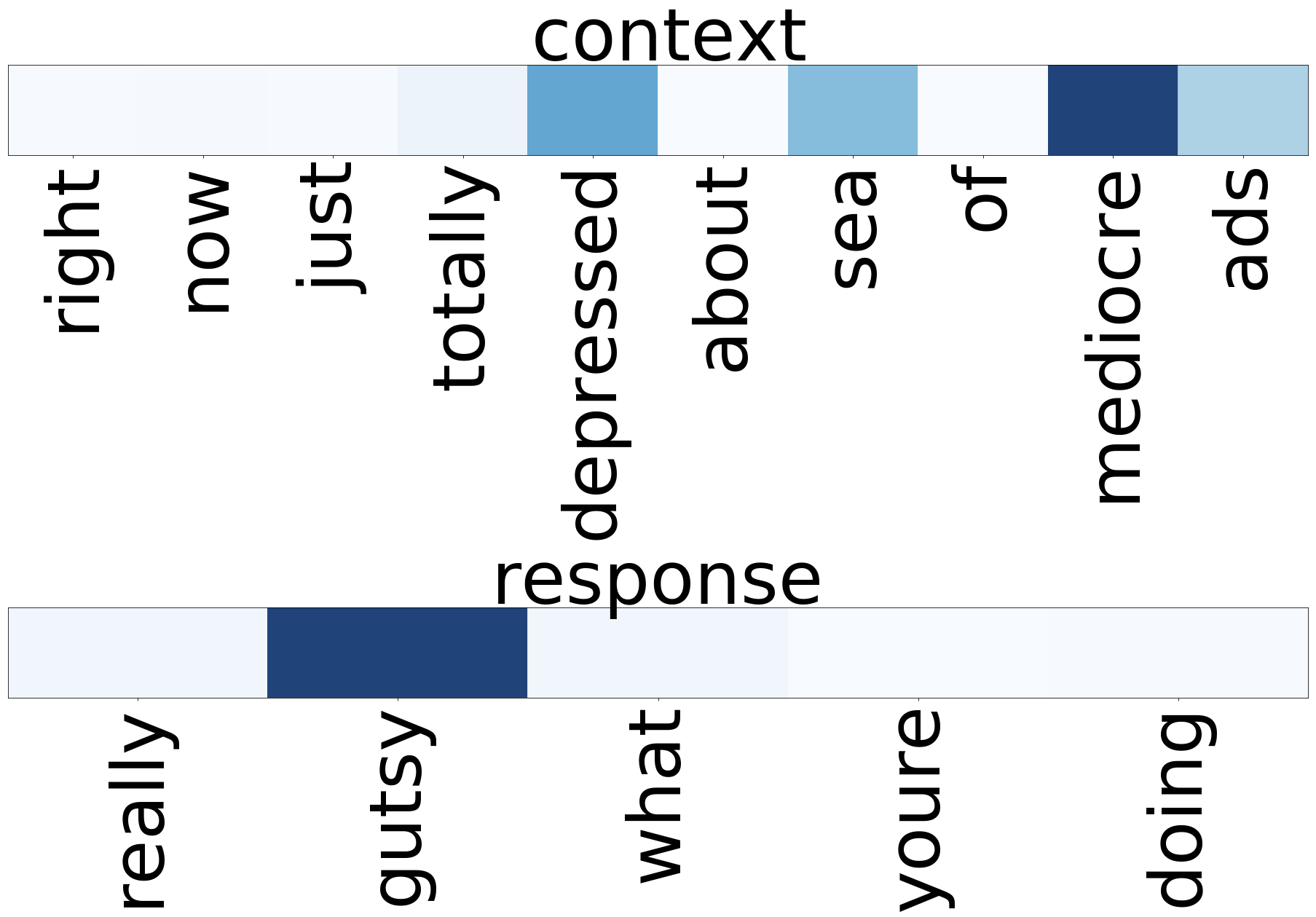}}
 \end{framed}
   \caption{Attention visualization of  incongruity between $c$ and $r$}
  \label{figure:wordmap2}
\end{figure}
\begin{figure}[t]
\centering
\begin{framed}
 {\includegraphics[width=1.7in]{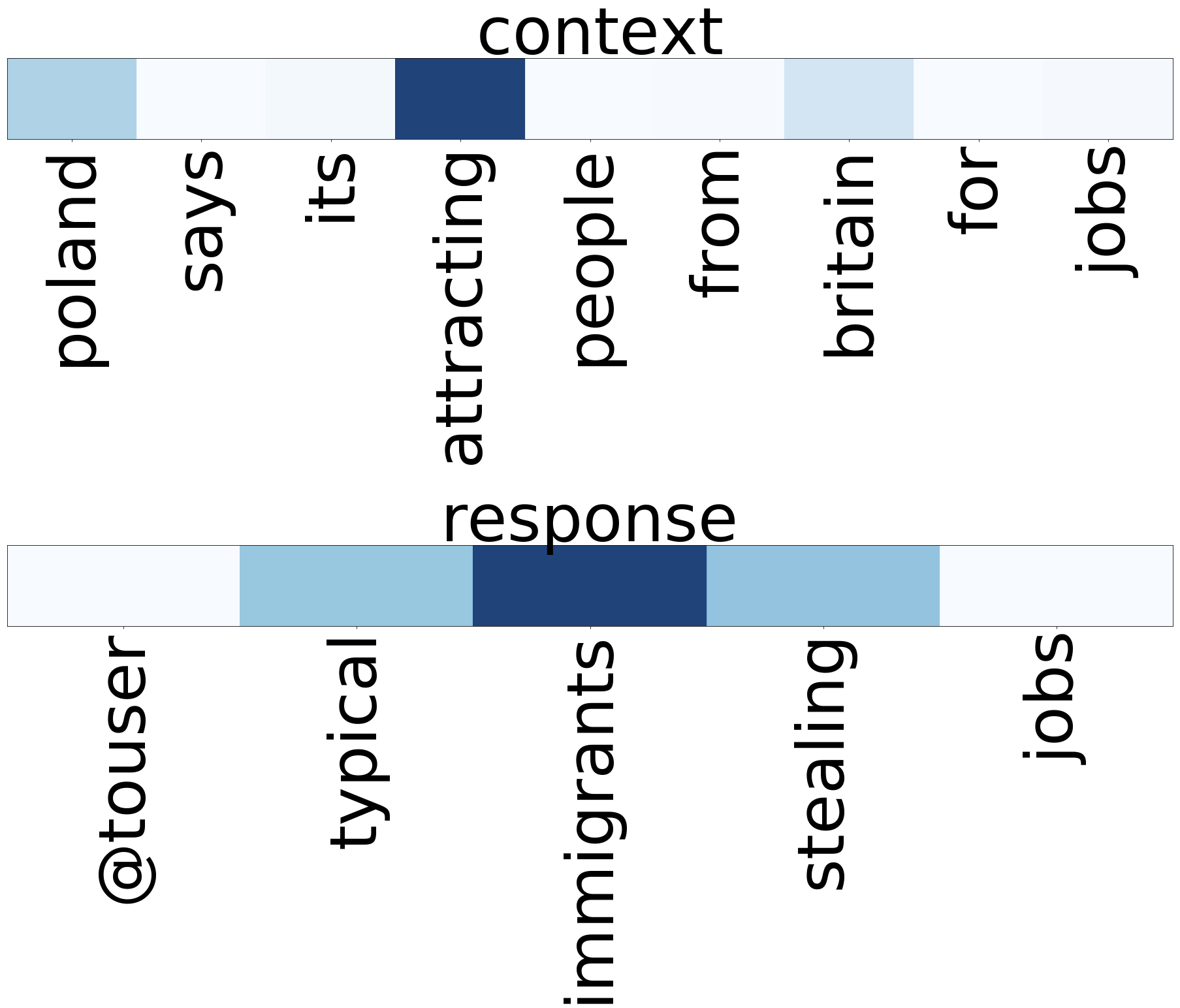}}
 \end{framed}
   \caption{Attention visualization of incongruity between $c$ and $r$ }
  \label{figure:wordmap3}
\end{figure}
\begin{figure}[t]
\centering
\begin{framed}
 {\includegraphics[width=1.7in]{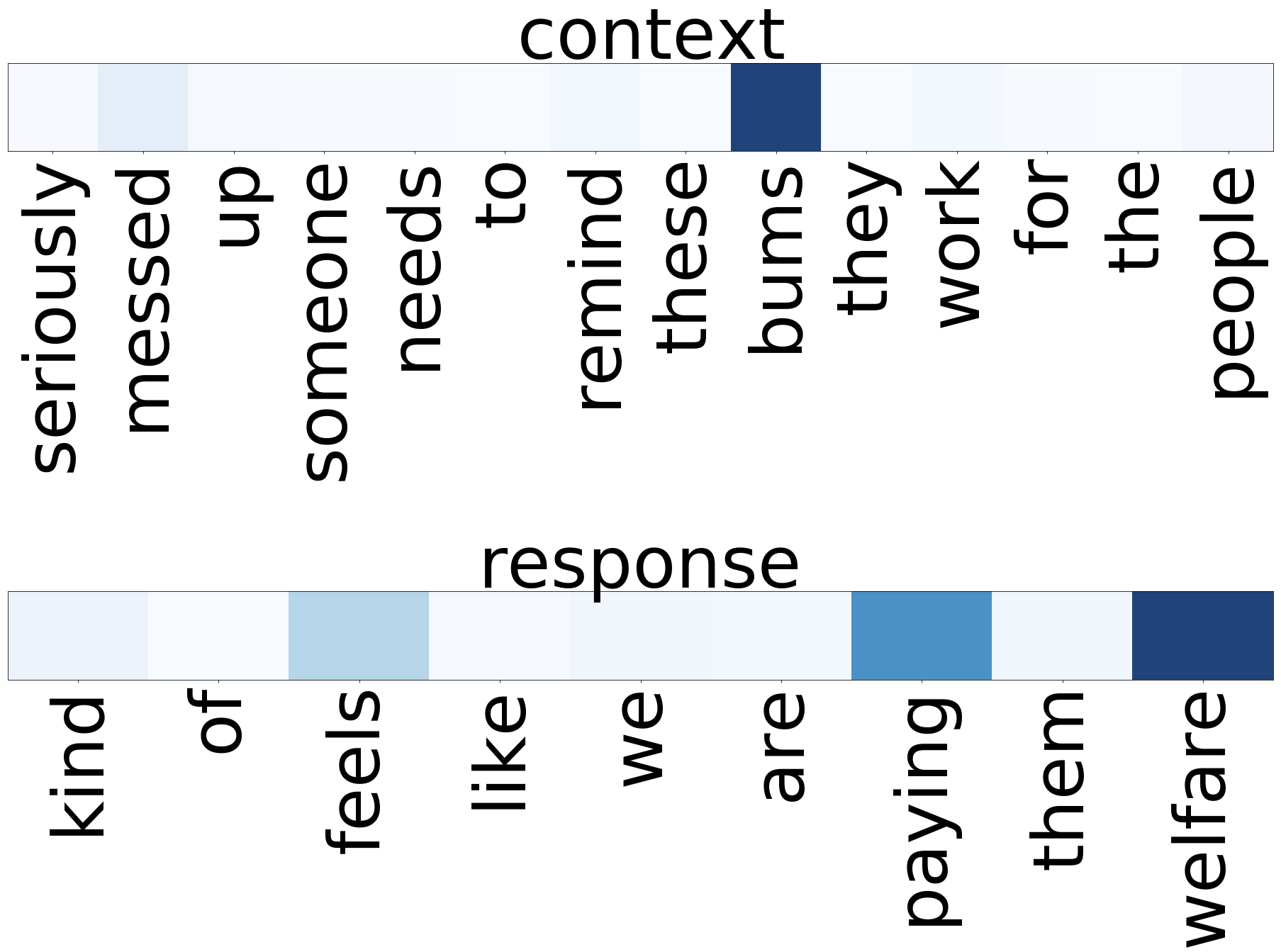}}
 \end{framed}
   \caption{Attention visualization of incongruity between $c$ and $r$ }
  \label{figure:wordmap4}
\end{figure}

In Twitter dataset, we observe often the attention models have selected utterance(s) from the context which have opposite sentiment (Figure \ref{figure:wordmap2}, Figure \ref{figure:wordmap3}, and Figure \ref{figure:wordmap4}). Here, the word and sentence-level attention model have chosen the particular utterance from the context (i.e., the top heatmap for the context) and the words with high attention (e.g., ``mediocre'', ``gutsy'').
These words again show examples of meaning incongruity which is useful for sarcasm detection. Word-models seem to also work well when words in the context/reply are semantically incongruous but connected via deeper semantics (``bums'' and ``welfare'' in context: ``someone needs to remind these \textit{bums} they work for the people'' and reply: ``feels like we are paying them \textit{welfare}'' (Figure \ref{figure:wordmap4}).  

\paragraph{Attention weights and sarcasm markers}
Looking just at attention weights in reply, we notice the models are giving highest weight to sentences that contain sarcasm markers, such as emoticons (i.e., ``:p'', ``:)'') and interjections (i.e., ``ah'', ``hmm''). Sarcasm markers are explicit indicators of sarcasm that signal that an utterance is sarcastic, such as the use of emoticons, uppercase spelling of words, or interjections. \cite{attardo2000irony,burgers2012verbal}.
Use of such markers in social media (particularly in Twitter) is extensive. 


While we have started to understand the semantic of attention weights in this task, more studies need to be carry out. \newcite{rocktaschel2015reasoning} have argued that interpretations based on attentions weights have to be taken with care since the classification task is not forced to solely rely on the attentions weights. Thus in future work, we plan to analyze utterances that are more subtle and do not consist of sarcasm markers or explicit incongruence of opposite sentiment between context and response.

\section{Related Work}

Most computational models for sarcasm detection have considered utterances in isolation  \cite{davidov2010,gonzalez,liebrecht2013perfect,riloff,maynard2014cares,ghoshguomuresan2015EMNLP,joshi2016word,ghosh2016fracking}. However, even humans have difficulty sometimes in recognizing sarcastic intent when considering an utterance in isolation \cite{wallace2014humans}. Thus, recent work on sarcasm and irony detection have started to exploit contextual information. In particular, \cite{khattri2015your} analyzed authors' prior sentiment towards certain entities and if a new tweet deviates from the author's estimated sentiment the tweet is predicted to be sarcastic. Similar to this approach, several models have been introduced; some relied on extensive feature engineering to capture contextual information about authors, topics or conversation context whereas the rest are using deep learning techniques to embed authors' information \cite{rajadesingan2015sarcasm}. The two studies that have considered conversation context among other contextual information have shown minimal improvement when modeling conversation context using Twitter data \cite{bamman2015contextualized,wang2015twitter}. Our work show that using better models, such as LSTM networks show a clear benefit of using context for sarcasm detection. As stated earlier in Section \ref{section:experiment}, LSTM's have been shown to be effective in NLI tasks, especially where the task is to establish the relationship between multiple inputs (i.e., in our case, between the context and the response). We observe that the LSTM$^{conditional}$ model and the sentence level attention-based models using both context and reply present the best results. 

\section{Conclusion}

This research makes a complementary contribution to existing work of modeling context for sarcasm/irony detection by looking at a particular type of context, \emph{conversation context}. We have addressed two issues: (1) does modeling of conversation context help in sarcasm detection
and (2) can we determine what part of the conversation context triggered the sarcastic reply. To answer the first question, we show that Long Short-Term Memory (LSTM) networks that can model both the  context and the sarcastic reply achieve better performance than LSTM networks that read only the reply. 
In particular, conditional LSTM networks \cite{rocktaschel2015reasoning} and LSTM networks with sentence level attention achieved significant improvement (e.g., 6-11\% F1 for discussion forums and Twitter messages). To address the second issue, we presented a qualitative analysis of attention weights produced by the LSTM models with attention, and discussed the results compared with human annotators. We also showed that attention-based models are able to identify inherent characteristics of sarcasm (i.e., sarcasm markers and sarcasm factors such as context incongruity). In future, we plan to study larger context, such as the full thread in a discussion forum that consider also the responses to the sarcastic comment, when available. We are also interested in analyzing sarcastic replies that do not contain sarcasm markers or explicit incongruence (i.e., opposing sentiment between the context and the reply).

\section*{Acknowledgements}
This paper is based on work supported by the DARPA-DEFT program. The views expressed are those of the authors and do not reflect the official policy or position of the Department of Defense or the U.S. Government. The authors thank Christopher Hidey for the discussions and resources on LSTM and the anonymous reviewers for helpful comments. 

\begin{thebibliography}{}
\expandafter\ifx\csname natexlab\endcsname\relax\def\natexlab#1{#1}\fi

\bibitem[{Attardo(2000)}]{attardo2000irony}
Salvatore Attardo. 2000.
\newblock Irony markers and functions: Towards a goal-oriented theory of irony
  and its processing.
\newblock {\em Rask\/} 12(1):3--20.

\bibitem[{Bamman and Smith(2015)}]{bamman2015contextualized}
David Bamman and Noah~A Smith. 2015.
\newblock Contextualized sarcasm detection on twitter.
\newblock In {\em Ninth International AAAI Conference on Web and Social
  Media\/}.

\bibitem[{Bird et~al.(2009)Bird, Klein, and Loper}]{bird2009natural}
Steven Bird, Ewan Klein, and Edward Loper. 2009.
\newblock {\em Natural language processing with Python: analyzing text with the
  natural language toolkit\/}.
\newblock " O'Reilly Media, Inc.".

\bibitem[{Bowman et~al.(2015)Bowman, Angeli, Potts, and
  Manning}]{bowman2015large}
Samuel~R Bowman, Gabor Angeli, Christopher Potts, and Christopher~D Manning.
  2015.
\newblock A large annotated corpus for learning natural language inference.
\newblock {\em arXiv preprint arXiv:1508.05326\/} .

\bibitem[{Burgers et~al.(2012)Burgers, Van~Mulken, and
  Schellens}]{burgers2012verbal}
Christian Burgers, Margot Van~Mulken, and Peter~Jan Schellens. 2012.
\newblock Verbal irony differences in usage across written genres.
\newblock {\em Journal of Language and Social Psychology\/} 31(3):290--310.

\bibitem[{Camp(2012)}]{camp2011}
Elisabeth Camp. 2012.
\newblock Sarcasm, pretense, and the semantics/pragmatics distinction*.
\newblock {\em No{\^u}s\/} 46(4):587--634.

\bibitem[{Chang and Lin(2011)}]{svm}
Chih-Chung Chang and Chih-Jen Lin. 2011.
\newblock {LIBSVM}: A library for support vector machines.
\newblock {\em ACM Transactions on Intelligent Systems and Technology\/}
  2:27:1--27:27.

\bibitem[{Davidov et~al.(2010)Davidov, Tsur, and Rappoport}]{davidov2010}
Dmitry Davidov, Oren Tsur, and Ari Rappoport. 2010.
\newblock Semi-supervised recognition of sarcastic sentences in twitter and
  amazon.
\newblock In {\em Proceedings of the Fourteenth Conference on Computational
  Natural Language Learning\/}. CoNLL '10.

\bibitem[{Ghosh and Veale(2016)}]{ghosh2016fracking}
Aniruddha Ghosh and Tony Veale. 2016.
\newblock Fracking sarcasm using neural network.
\newblock In {\em Proceedings of NAACL-HLT\/}. pages 161--169.

\bibitem[{Ghosh et~al.(2015)Ghosh, Guo, and Muresan}]{ghoshguomuresan2015EMNLP}
Debanjan Ghosh, Weiwei Guo, and Smaranda Muresan. 2015.
\newblock Sarcastic or not: Word embeddings to predict the literal or sarcastic
  meaning of words.
\newblock In {\em Proceedings of the 2015 Conference on Empirical Methods in
  Natural Language Processing\/}. Association for Computational Linguistics,
  Lisbon, Portugal, pages 1003--1012.

\bibitem[{Gimpel et~al.(2011)Gimpel, Schneider, O'Connor, Das, Mills,
  Eisenstein, Heilman, Yogatama, Flanigan, and Smith}]{gimpel2011part}
Kevin Gimpel, Nathan Schneider, Brendan O'Connor, Dipanjan Das, Daniel Mills,
  Jacob Eisenstein, Michael Heilman, Dani Yogatama, Jeffrey Flanigan, and
  Noah~A Smith. 2011.
\newblock Part-of-speech tagging for twitter: Annotation, features, and
  experiments.
\newblock In {\em Proceedings of the 49th Annual Meeting of the ACL\/}. pages
  42--47.

\bibitem[{Gonz{\'a}lez-Ib{\'a}{\~n}ez et~al.(2011)Gonz{\'a}lez-Ib{\'a}{\~n}ez,
  Muresan, and Wacholder}]{gonzalez}
Roberto Gonz{\'a}lez-Ib{\'a}{\~n}ez, Smaranda Muresan, and Nina Wacholder.
  2011.
\newblock Identifying sarcasm in twitter: A closer look.
\newblock In {\em ACL (Short Papers)\/}. Association for Computational
  Linguistics, pages 581--586.

\bibitem[{Grice et~al.(1975)Grice, Cole, and Morgan}]{grice1975syntax}
H~Paul Grice, Peter Cole, and Jerry~L Morgan. 1975.
\newblock Syntax and semantics.
\newblock {\em Logic and conversation\/} 3:41--58.

\bibitem[{Haverkate(1990)}]{haverkate1990speech}
Henk Haverkate. 1990.
\newblock A speech act analysis of irony.
\newblock {\em Journal of Pragmatics\/} 14(1):77--109.

\bibitem[{Hochreiter and Schmidhuber(1997)}]{hochreiter1997long}
Sepp Hochreiter and J{\"u}rgen Schmidhuber. 1997.
\newblock Long short-term memory.
\newblock {\em Neural computation\/} 9(8):1735--1780.

\bibitem[{Hu and Liu(2004)}]{hu2004mining}
Minqing Hu and Bing Liu. 2004.
\newblock Mining and summarizing customer reviews.
\newblock In {\em Proceedings of the tenth ACM SIGKDD international conference
  on Knowledge discovery and data mining\/}. ACM, pages 168--177.

\bibitem[{Joshi et~al.(2015)Joshi, Sharma, and Bhattacharyya}]{joshi2015}
Aditya Joshi, Vinita Sharma, and Pushpak Bhattacharyya. 2015.
\newblock Harnessing context incongruity for sarcasm detection.
\newblock In {\em Proceedings of the 53rd Annual Meeting of the Association for
  Computational Linguistics and the 7th International Joint Conference on
  Natural Language Processing (Volume 2: Short Papers)\/}. Association for
  Computational Linguistics, Beijing, China, pages 757--762.

\bibitem[{Joshi et~al.(2016)Joshi, Tripathi, Patel, Bhattacharyya, and
  Carman}]{joshi2016word}
Aditya Joshi, Vaibhav Tripathi, Kevin Patel, Pushpak Bhattacharyya, and Mark
  Carman. 2016.
\newblock Are word embedding-based features useful for sarcasm detection?
\newblock {\em arXiv preprint arXiv:1610.00883\/} .

\bibitem[{Khattri et~al.(2015)Khattri, Joshi, Bhattacharyya, and
  Carman}]{khattri2015your}
Anupam Khattri, Aditya Joshi, Pushpak Bhattacharyya, and Mark~James Carman.
  2015.
\newblock Your sentiment precedes you: Using an author’s historical tweets to
  predict sarcasm.
\newblock In {\em 6th Workshop on Computational Approaches to Subjectivity,
  Sentiment and Social Media Analysis (WASSA)\/}. page~25.

\bibitem[{Liebrecht et~al.(2013)Liebrecht, Kunneman, and van~den
  Bosch}]{liebrecht2013perfect}
CC~Liebrecht, FA~Kunneman, and APJ van~den Bosch. 2013.
\newblock The perfect solution for detecting sarcasm in tweets\# not .

\bibitem[{Maynard and Greenwood(2014)}]{maynard2014cares}
Diana Maynard and Mark~A Greenwood. 2014.
\newblock Who cares about sarcastic tweets? investigating the impact of sarcasm
  on sentiment analysis.
\newblock In {\em Proceedings of LREC\/}.

\bibitem[{Mikolov et~al.(2013)Mikolov, Chen, Corrado, and
  Dean}]{mikolov2013efficient}
Tomas Mikolov, Kai Chen, Greg Corrado, and Jeffrey Dean. 2013.
\newblock Efficient estimation of word representations in vector space.
\newblock {\em arXiv preprint arXiv:1301.3781\/} .

\bibitem[{Muresan et~al.(2016)Muresan, Gonzalez-Ibanez, Ghosh, and
  Wacholder}]{muresanjasist2016}
Smaranda Muresan, Roberto Gonzalez-Ibanez, Debanjan Ghosh, and Nina Wacholder.
  2016.
\newblock Identification of nonliteral language in social media: A case study
  on sarcasm.
\newblock {\em Journal of the Association for Information Science and
  Technology\/} .

\bibitem[{Oraby et~al.(2016)Oraby, Harrison, Hernandez, Reed, Riloff, and
  Walker}]{orabycreating}
Shereen Oraby, Vrindavan Harrison, Ernesto Hernandez, Lena Reed, Ellen Riloff,
  and Marilyn Walker. 2016.
\newblock Creating and characterizing a diverse corpus of sarcasm in dialogue .

\bibitem[{Parikh et~al.(2016)Parikh, T{\"a}ckstr{\"o}m, Das, and
  Uszkoreit}]{parikh2016decomposable}
Ankur~P Parikh, Oscar T{\"a}ckstr{\"o}m, Dipanjan Das, and Jakob Uszkoreit.
  2016.
\newblock A decomposable attention model for natural language inference.
\newblock {\em arXiv preprint arXiv:1606.01933\/} .

\bibitem[{Pennebaker et~al.(2001)Pennebaker, Francis, and
  Booth}]{pennebaker2001}
James~W Pennebaker, Martha~E Francis, and Roger~J Booth. 2001.
\newblock Linguistic inquiry and word count: Liwc 2001.
\newblock {\em Mahway: Lawrence Erlbaum Associates\/} 71:2001.

\bibitem[{Rajadesingan et~al.(2015)Rajadesingan, Zafarani, and
  Liu}]{rajadesingan2015sarcasm}
Ashwin Rajadesingan, Reza Zafarani, and Huan Liu. 2015.
\newblock Sarcasm detection on twitter: A behavioral modeling approach.
\newblock In {\em Proceedings of the Eighth ACM International Conference on Web
  Search and Data Mining\/}. ACM, pages 97--106.

\bibitem[{Riloff et~al.(2013)Riloff, Qadir, Surve, De~Silva, Gilbert, and
  Huang}]{riloff}
Ellen Riloff, Ashequl Qadir, Prafulla Surve, Lalindra De~Silva, Nathan Gilbert,
  and Ruihong Huang. 2013.
\newblock Sarcasm as contrast between a positive sentiment and negative
  situation.
\newblock In {\em Proceedings of the Conference on Empirical Methods in Natural
  Language Processing\/}. Association for Computational Linguistics, pages
  704--714.

\bibitem[{Rockt{\"a}schel et~al.(2015)Rockt{\"a}schel, Grefenstette, Hermann,
  Ko{\v{c}}isk{\`y}, and Blunsom}]{rocktaschel2015reasoning}
Tim Rockt{\"a}schel, Edward Grefenstette, Karl~Moritz Hermann, Tom{\'a}{\v{s}}
  Ko{\v{c}}isk{\`y}, and Phil Blunsom. 2015.
\newblock Reasoning about entailment with neural attention.
\newblock {\em arXiv preprint arXiv:1509.06664\/} .

\bibitem[{Sutskever et~al.(2014)Sutskever, Vinyals, and
  Le}]{sutskever2014sequence}
Ilya Sutskever, Oriol Vinyals, and Quoc~V Le. 2014.
\newblock Sequence to sequence learning with neural networks.
\newblock In {\em Advances in neural information processing systems\/}. pages
  3104--3112.

\bibitem[{Tchokni et~al.(2014)Tchokni, S{\'e}aghdha, and Quercia}]{tchokni2014}
Simo Tchokni, Diarmuid~O S{\'e}aghdha, and Daniele Quercia. 2014.
\newblock Emoticons and phrases: Status symbols in social media.
\newblock In {\em Eighth International AAAI Conference on Weblogs and Social
  Media\/}.

\bibitem[{Vinyals et~al.(2015)Vinyals, Kaiser, Koo, Petrov, Sutskever, and
  Hinton}]{vinyals2015grammar}
Oriol Vinyals, {\L}ukasz Kaiser, Terry Koo, Slav Petrov, Ilya Sutskever, and
  Geoffrey Hinton. 2015.
\newblock Grammar as a foreign language.
\newblock In {\em Advances in Neural Information Processing Systems\/}. pages
  2773--2781.

\bibitem[{Wallace et~al.(2014)Wallace, Do~Kook~Choe, Kertz, and
  Charniak}]{wallace2014humans}
Byron~C Wallace, Laura~Kertz Do~Kook~Choe, Laura Kertz, and Eugene Charniak.
  2014.
\newblock Humans require context to infer ironic intent (so computers probably
  do, too).
\newblock In {\em ACL (2)\/}. pages 512--516.

\bibitem[{Wang et~al.(2015)Wang, Wu, Wang, and Ren}]{wang2015twitter}
Zelin Wang, Zhijian Wu, Ruimin Wang, and Yafeng Ren. 2015.
\newblock Twitter sarcasm detection exploiting a context-based model.
\newblock In {\em International Conference on Web Information Systems
  Engineering\/}. Springer, pages 77--91.

\bibitem[{Wilson et~al.(2005)Wilson, Wiebe, and
  Hoffmann}]{wilson2005recognizing}
Theresa Wilson, Janyce Wiebe, and Paul Hoffmann. 2005.
\newblock Recognizing contextual polarity in phrase-level sentiment analysis.
\newblock In {\em Proceedings of the conference on human language technology
  and empirical methods in natural language processing\/}. Association for
  Computational Linguistics, pages 347--354.

\bibitem[{Xu et~al.(2015)Xu, Ba, Kiros, Cho, Courville, Salakhudinov, Zemel,
  and Bengio}]{xu2015show}
Kelvin Xu, Jimmy Ba, Ryan Kiros, Kyunghyun Cho, Aaron Courville, Ruslan
  Salakhudinov, Rich Zemel, and Yoshua Bengio. 2015.
\newblock Show, attend and tell: Neural image caption generation with visual
  attention.
\newblock In {\em International Conference on Machine Learning\/}. pages
  2048--2057.

\bibitem[{Yang et~al.(2016)Yang, Yang, Dyer, He, Smola, and
  Hovy}]{yang2016hierarchical}
Zichao Yang, Diyi Yang, Chris Dyer, Xiaodong He, Alex Smola, and Eduard Hovy.
  2016.
\newblock Hierarchical attention networks for document classification.
\newblock In {\em Proceedings of NAACL-HLT\/}. pages 1480--1489.

\bibitem[{Yin et~al.(2015)Yin, Sch{\"u}tze, Xiang, and Zhou}]{yin2015abcnn}
Wenpeng Yin, Hinrich Sch{\"u}tze, Bing Xiang, and Bowen Zhou. 2015.
\newblock Abcnn: Attention-based convolutional neural network for modeling
  sentence pairs.
\newblock {\em arXiv preprint arXiv:1512.05193\/} .

\bibitem[{Zaremba et~al.(2014)Zaremba, Sutskever, and
  Vinyals}]{zaremba2014recurrent}
Wojciech Zaremba, Ilya Sutskever, and Oriol Vinyals. 2014.
\newblock Recurrent neural network regularization.
\newblock {\em arXiv preprint arXiv:1409.2329\/} .

\end{thebibliography}

\bibliographystyle{acl_natbib}

\end{document}